\documentclass[12pt,a4paper,twoside]{article}

\usepackage[a4paper,top=20mm,bottom=20mm,inner=25mm,outer=25mm]{geometry}
\usepackage[T1]{fontenc}
\usepackage[utf8]{inputenc}
\usepackage[sfdefault]{arimo}

\usepackage[english]{babel}
\usepackage{microtype}
\usepackage{ragged2e}
\usepackage{amsmath,amssymb}

\usepackage{graphicx}
\usepackage{array}
\usepackage{multirow}
\usepackage{booktabs}
\usepackage{caption}
\usepackage{tcolorbox}
\usepackage{pgfplots}
\pgfplotsset{compat=1.18}
\usepackage{titlesec}
\titleformat{\section}{\normalfont\bfseries\fontsize{12}{14}\selectfont}{}{0pt}{}
\titleformat{\subsection}{\normalfont\bfseries\itshape\fontsize{12}{14}\selectfont}{}{0pt}{}
\titleformat{\subsubsection}{\normalfont\itshape\fontsize{12}{14}\selectfont}{}{0pt}{}
\titlespacing*{\section}{0pt}{0.8em}{0.35em}
\titlespacing*{\subsection}{0pt}{0.5em}{0.25em}
\titlespacing*{\subsubsection}{0pt}{0.4em}{0.15em}

\usepackage[backend=biber,style=apa,natbib=true,sorting=nyt]{biblatex}
\DeclareLanguageMapping{english}{english-apa}
\newif\ifreview

\newcommand{\sefiincludegraphics}[2][]{%
  \IfFileExists{#2}{\includegraphics[#1]{#2}}{%
    \fbox{\parbox[c][45mm][c]{0.9\linewidth}{\centering Missing figure file: \texttt{#2}}}%
  }%
}

\usepackage{fancyhdr}

\fancypagestyle{firstpage}{
    \fancyhf{}
    \fancyhead[L]{\small\textit{SEFI 2026: 54th Annual Conference of the European Society for Engineering Education (SEFI)}}
    
}

\begin{document}

\thispagestyle{firstpage}

\begin{center}
    {\bfseries\fontsize{14}{16}\selectfont An Evaluation of AI-Supported Evidence-Based Learning for Public Speaking Skill Development\par}
\end{center}

\vspace{0.8em}

\ifreview
    \begin{center}
        \vspace{2.5\baselineskip}
        {\itshape Anonymous Author(s)}\par
        \vspace{2.5\baselineskip}
    \end{center}
\else
    \begin{center}
        \textbf{Sashini Hettiarachchi}, \textbf{Shahbaz Siddeeq\kern-0.15em\renewcommand{\thefootnote}{\fnsymbol{footnote}}
\footnote{Corresponding author: Shahbaz Siddeeq  \textless\textit{shahbaz.siddeeq@tuni.fi}\textgreater}
\renewcommand{\thefootnote}{\arabic{footnote}}}, \textbf{Mika Saari}, \textbf{Pekka Abrahamsson}\\
        \vspace{1\baselineskip}
    	\normalsize Computing Sciences, Tampere University, Pori, Finland\\

        
    \end{center}
    
\fi

\textit{\textbf{Conference Key Areas:} Engineering skills, professional skills, and transversal skills;  Digital tools and AI in engineering education}


\section*{Abstract}
Public speaking is an essential skill in academic and professional contexts, but it often causes anxiety. Although several AI-based speech coaching tools exist, they typically provide generic feedback and lack model speeches for learning. This study addresses these gaps by developing a system that incorporates speech context into evaluation criteria, enabling more tailored feedback. It also generates improved model speeches in both text and audio formats to support example-based learning. 
The system was evaluated in a pilot study with seven undergraduate students over one to three weeks. Participants used the system repeatedly and practiced the same speech at least three times. Results showed improvements in public speaking performance for all participants. Filler word usage decreased, and anxiety levels dropped by 15.3\% to 34.4\%. Participants found the context-aware feedback and revised speeches useful and confidence-building.
These findings suggest that context-aware feedback and model speeches can enhance public speaking skills and reduce anxiety. Future studies could explore video-based analysis of physical behavior during speeches.

\section{Introduction}

Public speaking is a crucial skill in academic and professional development. However, public speaking is one of the most anxiety provoking tasks for many people. Therefore, people frequently experiment with various approaches to improving their public speaking skills, whether it be enrolling in public speaking classes, joining organizations like Toastmasters \parencite{toastmasters}, or attending mentoring sessions \parencite{dwyer2021take}. These methods give the opportunity to practice speeches, receive feedback, and improve. However, these traditional methods can be limited by time, instructor availability, and the subjective nature of feedback \parencite{mohd2023factors,isotalus2024artificial}.

The development of artificial intelligence (AI) has opened new opportunities to make public speaking feedback more personalized and accessible. Advancements in speech recognition using Speech-to-Text (STT) technology, natural language processing (NLP) with Language Models (LM) and prosody analysis enable AI tools to understand and evaluate speech patterns, detect filler words, measure vocal variety \parencite{padia2024enhancing}, and even generate improved versions of speech content. Despite that, there are limitations in existing AI-based public speaking applications. First, most current AI speech coaching applications provide generic feedback that overlooks the purpose of the speech, audience, or context \parencite{isotalus2024artificial}. This makes the feedback less meaningful for real-world speaking situations. Second, most of the time, these systems do not offer example speeches, although model examples are an important part of learning. The VoiceCoach application is one of the best examples of research that shows the impact of example-based learning \parencite{Wangetal2020}. It demonstrates the effectiveness of speech training by recommending examples of vocal modulation from experts' speeches for learners.

This study addresses these gaps by developing a Context-Aware AI Speech Coach application. The Context-Aware AI Speech Coach is a web application that evaluate user's speeches in a standard method  by using the Competent Speaker Speech Evaluation Framework (CSSEF) \parencite{nca2007competent}. In addition, this system generates an improved version of the user’s speech as a model speech in both text and audio formats.  For evaluation and model speech generation, the system uses contextual information about the speech, such as its communicative goal and audience. This will fill gaps in previous research that did not use contextual information in feedback generation and lacked model examples for example-based learning. In doing so, the system adopts an evidence-based learning approach, in which improvement is guided by concrete model speeches and feedback grounded in an established speech evaluation framework (CSSEF) rather than generic advice. Therefore, the aim of this research is to examine whether context-aware AI feedback and revised speech examples can improve public speaking performance and reduce speaking anxiety. Specifically, the study seeks to: (1) develop a context-aware AI speech coaching system grounded in the CSSEF framework; (2) provide improved example speeches in text and audio as learning models; and (3) evaluate the system’s impact on speaking performance, learner reflection, and public speaking anxiety.

Accordingly, the study investigates the following research questions:

\textbf{RQ1:} How effective is context-aware AI-generated feedback for improving public speaking skills and perceived feedback quality?


\textbf{RQ2:} How do AI-generated revised speeches (text and audio) improve public speaking and learners’ perceptions of their performance?

\textbf{RQ3:} To what extent does using the AI Speech Coach reduce learners’ public speaking anxiety?
 

\section{Literature review}

Recent AI research highlights that most AI systems lack intent and epistemic awareness. This means that AI does not understand why a speaker is speaking unless context is explicitly provided. The Harvard Misinfo Review explains that AI performs unevenly across tasks because it lacks intent awareness. This is known as artificial jagged intelligence \parencite{flew2024new}. Therefore, this study focuses on providing AI feedback that is based on user-provided context.




The National Communication Association (NCA) developed the Competent Speaker Speech Evaluation Form (CSSEF) \parencite{nca2007competent} to provide a standardized and impartial way to assess a speech. The CSSEF assesses eight competencies that cover both the planning and delivery of a speech. The rating system to evaluate each competency is inadequate, satisfactory, or exceptional \parencite{nca2007competent}.

\textcite{garcia2024addressing} evaluated improvements in students’ public speaking performance using CSSEF, where an English teacher evaluated the pre- and post-experiment speeches. Therefore, to compare the results gained in the previous study, CSSEF was selected for this study. However, the current study evaluation was done by AI. Further, according to \textcite{garcia2024addressing} and the National Communication Association (NCA) \parencite{nca2007competent}, CSSEF is frequently used to evaluate public speaking proficiency at the higher education level.


Artificial Intelligence (AI) has become a transformative tool in communication education particularly in developing public speaking skills. AI powered speech coaching platforms now provide learners with personalized, data-driven feedback and the opportunity to learn independently, anywhere and anytime. However, despite their growing sophistication, most of the AI public speaking practicing systems still lack context awareness of the speeches \parencite{isotalus2024artificial}.

Even though the AI platforms have advanced the efficiency and accessibility of learning public speaking, a critical limitation of lack of context awareness still exists. Most existing AI public speaking training tools evaluate the audio or video of the speech independently, focusing on surface-level delivery metrics rather than contextual variables such as target audience, goal of the speech, and type of speech. The study by \textcite{isotalus2024artificial} conducted with MySpeaker Rhetorich \parencite{myspeaker_rhetorich} highlights that AI could assess paralinguistic elements such as pacing and pitch, but failed to consider the subject, context, or target audience of the speech.



Revised versions of speeches in audio, video, or textual form play an important role in improving public speaking skills. \textcite{budiasningrum2022using} showed that repeated exposure to audio materials, such as podcasts, strengthens listening skills and improves pronunciation and fluency. Similarly, the VoiceCoach application illustrates the value of model examples by enhancing voice training outcomes through the recommendation of high quality vocal modulation samples \parencite{Wangetal2020}. However, \textcite{garcia2024addressing} highlights a limitation of existing tools, as it primarily addresses performance anxiety while offering limited support for process anxiety, particularly in guiding learners on how to construct and revise a speech.

Previous work by \textcite{mohd2023factors} observed that postgraduate students who received kinder and more supportive feedback improved their confidence and reduced their PSA over time. Therefore, the feedback mechanism serves as a bridge between confidence building, learning, and reducing the fear of public speaking.

Previous research by \textcite{BA2025100284} showed that AI can effectively enhance targeted learning outcomes by providing feedback tailored to different aspects of student performance. Further, \textcite{kalisa-compare-ai-human-feedback} showed that AI-generated feedback is as effective as human-generated feedback in supporting learning outcomes.

This highlights the importance of feedback and the potential of AI as an effective feedback provider in education and learning.

Practicing also plays an important role in reducing public speaking anxiety (PSA), as repeated exposure and preparation beforehand help speakers build confidence and familiarity with their delivery. \textcite{mohd2023factors} reported that one of the primary strategies students shared for mitigating PSA was preparation before the presentation. \textcite{dwyer2021take} found that public speaking courses can significantly help reduce fear of public speaking and build students' confidence.

\section{Methodology}







\subsection{Overview of the experiment}

The experiment was conducted over a limited period with seven participants. Participants were required to interact with AI speech coach applications multiple times and practice the same speech at least three times. While some participants completed three practice sessions within one week, others required up to three weeks to complete the experiment. The study was conducted alongside the regular academic commitments of the participants, which limited the amount of time they could dedicate exclusively to speech practice.
All participants were undergraduate students at the University of Moratuwa, Sri Lanka, aged between 21 and 23 years, including three female and four male students. English was a second language for all participants. To ensure anonymity, each participant was assigned a unique identifier, labeled P1 to P7.

\subsection{Speech Analysis Process}

\begin{figure}[thbp]
  \centering
  \sefiincludegraphics[width=\textwidth,clip, trim=20 390 200 35]{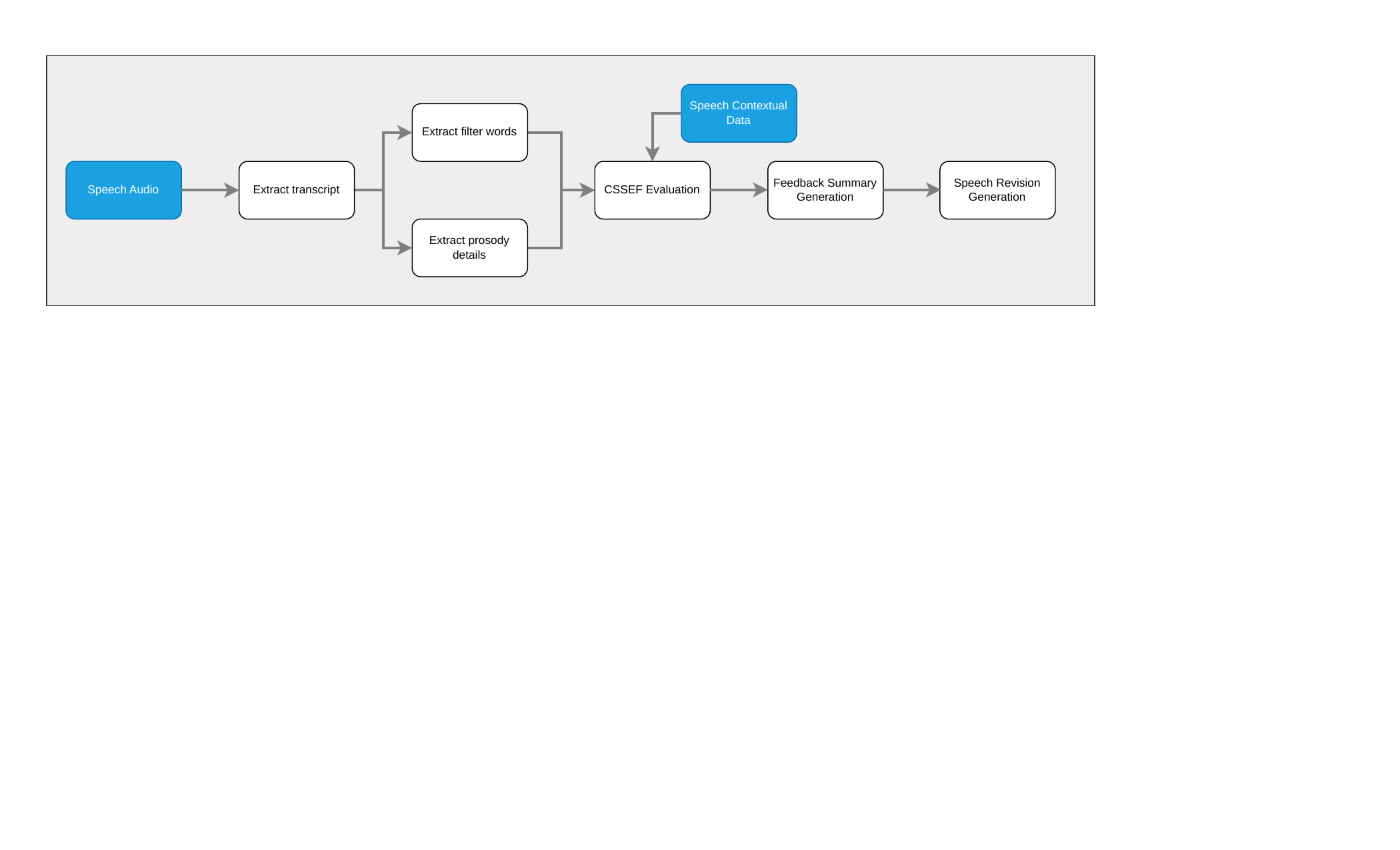}
  \caption{Speech Analysis Pipeline}
  \label{fig:speech-analysing}
\end{figure}

As shown in Figure \ref{fig:speech-analysing}, all speeches were uploaded to the system as audio files and processed through the AI speech coaching pipeline. First, a transcript was generated from the audio uploaded. The system then identified filler words such as “um,” “uh,” and “like,” and calculated their frequency relative to the total word count. Prosodic features, including pitch variation, speech rate, pauses, and volume changes, were also extracted. These features have been widely used in previous studies to assess public speaking competence \parencite{padia2024enhancing,garcia2024addressing,isotalus2024artificial}.
Each speech was evaluated against seven competencies of the Competent Speaker Speech Evaluation Form (CSSEF) using the GPT-4o model. The eighth competency, Physical Behaviors (C8), was excluded because the current system implementation supports audio and text analysis only. Each competency was scored on a 1–5 scale, adapted from the CSSEF's original three-point (inadequate, satisfactory, exceptional) scale to provide finer granularity. Scores of 4 and above were classified as Excellent, scores from 2 up to (but not including) 4 as Satisfactory, and scores below 2 as Unsatisfactory. The overall score reported for each session was computed as the sum of the competency scores expressed as a percentage of the maximum achievable score. The system generated a concise summary that included an overall score, key strengths, and areas for improvement.


After each submission, the AI system generated two enhanced versions of the speech. The revised text version focused on improving structure, clarity, organization, and flow by incorporating contextual information provided by the user and the CSSEF evaluation results. In addition, an audio version was generated based on the revised text. This audio served as a model example, demonstrating clearer articulation, appropriate pacing, increased vocal variety, and stronger delivery.


\subsection{Experimental Procedure}

Participants first completed a pre-experiment survey that included informed consent and background questions. They then attended a 30 minute online orientation session, which included a demonstration of the AI speech coaching system. After logging into the platform, participants completed the Personal Report of Public Speaking Anxiety (PRPSA) questionnaire to establish a baseline measure of public speaking anxiety. The PRPSA is a validated instrument consisting of 34 items rated on a five-point Likert scale, with total scores ranging from 34 to 170, where higher scores indicate greater anxiety \parencite{mccroskey1970effect}.

Participants then selected an academically relevant topic aligned with their interests. Each speech was required to include a clear title, a specific purpose, an intended audience, and key points, with a target duration of 2–4 minutes.
Participants engaged in an iterative learning process by reviewing the AI generated feedback and revised speech materials. They were instructed to read the improved text version, listen to the revised audio version, reflect on differences between their original and improved performances, and revise and re-record the same speech by applying the feedback.


After completing all practice sessions, participants completed the PRPSA again through the system to assess changes in public speaking anxiety. Finally, participants responded to a post-experiment questionnaire that examined their perceptions of the feedback quality, the usefulness of the revised speeches, and their perceived changes in nervousness and anxiety after using the system.

\section{Results and analysis}
\subsection{Effectiveness of Context Aware AI Generated Feedback}

\begin{table}[h]
\centering
\caption{CSSEF Overall Score Changes}
\label{tab:overrall-cssef-changes}
\resizebox{\columnwidth}{!}{%
\begin{tabular}{|c|c|c|c|c|}
\hline
\textbf{Participant} &
\textbf{Sessions} &
\textbf{\begin{tabular}[c]{@{}c@{}}Initial Overall\\ Score (\%)\end{tabular}} &
\textbf{\begin{tabular}[c]{@{}c@{}}Final Overall\\ Score (\%)\end{tabular}} &
\textbf{\begin{tabular}[c]{@{}c@{}}Relative\\ Change (\%)\end{tabular}} \\
\hline
P1 & 4 & 57 & 67 & +17.5\% \\ \hline
P2 & 5 & 46 & 60 & +30.4\% \\ \hline
P3 & 4 & 62 & 66 & +6.5\%  \\ \hline
P4 & 3 & 62 & 72 & +16.1\% \\ \hline
P5 & 4 & 30 & 66 & +120.0\% \\ \hline
P6 & 5 & 57 & 63 & +10.5\% \\ \hline
P7 & 3 & 44 & 59 & +34.1\% \\ \hline
\end{tabular}
}
\end{table}


To evaluate an objective improvement in public speaking performance, CSSEF overall scores from the first and final context-based academic speech sessions were compared, as shown in Table \ref{tab:overrall-cssef-changes}. P2, P5, and P7 showed substantial and consistent gains across sessions, and P5 improved across all seven competencies, demonstrating the largest relative gain of +120\%.

Participants with moderate PSA demonstrated a relative improvement (+28.8\%) in CSSEF scores between the first and final academic speech sessions. The participant in the low PSA group also showed meaningful improvement (+16.1\%), though the magnitude was smaller. No participants were classified as high PSA at baseline, which limits direct comparison for this category.
Compared to the previous research done with Yoodli \parencite{garcia2024addressing}, where CSSEF improvements ranged between 54--70\% over a structured multi-day intervention, the present study demonstrates comparable proportional gains, particularly for participants with moderate PSA levels. While absolute gains are smaller due to fewer sessions and a smaller sample size, the direction and consistency of improvement align closely with prior findings, supporting the effectiveness of context-aware feedback even in a short term intervention.

\subsection{Impact of Revised Speeches}

\begin{table}[h]
\centering
\caption{Change in Filler Word Usage by Participant}
\label{tab:changes-in-filler-words}
\resizebox{\columnwidth}{!}{%
\begin{tabular}{|c|c|c|c|c|c|}
\hline
\textbf{Participant} &
\textbf{Sessions} &
\textbf{\begin{tabular}[c]{@{}c@{}}Initial Filler\\ Words (\%)\end{tabular}} &
\textbf{\begin{tabular}[c]{@{}c@{}}Final Filler\\ Words (\%)\end{tabular}} &
\textbf{\begin{tabular}[c]{@{}c@{}}Absolute\\ Change (pp)\end{tabular}} &
\textbf{\begin{tabular}[c]{@{}c@{}}Relative\\ Change (\%)\end{tabular}} \\
\hline
P1 & 4 & 2.4\% & 1.0\% & $-1.4$ & $-58.3$\% \\ \hline
P2 & 5 & 0.7\% & 0.0\% & $-0.7$ & $-100.0$\% \\ \hline
P3 & 4 & 0.7\% & 0.0\% & $-0.7$ & $-100.0$\% \\ \hline
P4 & 3 & 3.4\% & 1.3\% & $-2.1$ & $-61.8$\% \\ \hline
P5 & 4 & 6.3\% & 1.7\% & $-4.6$ & $-73.0$\% \\ \hline
P6 & 5 & 4.0\% & 2.7\% & $-1.3$ & $-32.5$\% \\ \hline
P7 & 3 & 2.0\% & 1.5\% & $-0.5$ & $-25.0$\% \\ \hline
\end{tabular}
}
\end{table}

\begin{table}[h]
\centering
\caption{Participants Ratings on Usefulness of Revised Speeches}
\label{tab:particpants-rating-revised-speech}
\resizebox{\columnwidth}{!}{%
\begin{tabular}{|p{9.5cm}|c|}
\hline
\textbf{Survey Question} &
\textbf{Average Rating} \\
\hline
The revised text version helped me see how to improve my structure.
& 4.71 \\ \hline

The AI-generated audio version helped me understand pacing and tone.
& 4.14 \\ \hline

Comparing my speech with the AI version improved my reflection.
& 4.14 \\ \hline

The improved versions increased my confidence to try again.
& 4.57 \\ \hline

I learned more effectively by listening to the AI-generated examples.
& 4.43 \\ \hline

Overall, the improved versions accelerated my learning.
& 4.71 \\ \hline
\end{tabular}
}
\end{table}

CSSEF Criterion 4 (Organization) showed consistent positive change across participants.
Six out of seven participants demonstrated improvement in this competency, with gains
ranging from +1 to +2. CSSEF Criterion 6 (Vocal Variety) also demonstrated notable improvement across participants. Most participants showed positive gains (+1), indicating enhanced control over pitch variation, pacing, and emphasis. These improvements align with participants’ use of AI generated audio versions, which provided concrete examples of effective delivery.
The table \ref{tab:changes-in-filler-words}  shows a reduction in filler word usage for all participants, with relative decreases ranging from 25\% to 100\%. Reduced filler word usage is commonly associated with increased confidence and improved speech planning, reinforcing the conclusion that revised speech versions supported smoother and more controlled delivery.

Table \ref{tab:particpants-rating-revised-speech} shows the Participants ratings on usefulness of AI generated revised speeches (5 point Likert scale: 1 = Strongly Disagree, 5 = Strongly Agree; n = 7). All the questions got the rating more than 4 out of 5. 

\subsection{Effect of AI Speech Coaching Application on Public Speaking Anxiety}

\begin{table}[h]
\centering
\caption{Individual PRPSA Score (Pre vs Post Intervention)}
\label{tab:invidual-prpsa-score-changes}
\resizebox{\columnwidth}{!}{%
\begin{tabular}{|c|c|c|c|c|c|c|}
\hline
\textbf{Participant} &
\textbf{\begin{tabular}[c]{@{}c@{}}Initial\\ PRPSA\end{tabular}} &
\textbf{\begin{tabular}[c]{@{}c@{}}PSA Level\\ (Pre)\end{tabular}} &
\textbf{\begin{tabular}[c]{@{}c@{}}Final\\ PRPSA\end{tabular}} &
\textbf{\begin{tabular}[c]{@{}c@{}}PSA Level\\ (Post)\end{tabular}} &
\textbf{\begin{tabular}[c]{@{}c@{}}Absolute\\ Change\end{tabular}} &
\textbf{\begin{tabular}[c]{@{}c@{}}Percentage\\ Change (\%)\end{tabular}} \\
\hline
P1 & 115 & Moderate & 77  & Low      & $-38$ & $-33.0$\% \\ \hline
P2 & 101 & Moderate & 67  & Low      & $-34$ & $-33.7$\% \\ \hline
P3 & 131 & Moderate & 86  & Low      & $-45$ & $-34.4$\% \\ \hline
P4 & 86  & Low      & 61  & Low      & $-25$ & $-29.1$\% \\ \hline
P5 & 126 & Moderate & 87  & Low      & $-39$ & $-31.0$\% \\ \hline
P6 & 124 & Moderate & 105 & Moderate & $-19$ & $-15.3$\% \\ \hline
P7 & 102 & Moderate & 76  & Low      & $-26$ & $-25.5$\% \\ \hline
\end{tabular}
}
\end{table}


\begin{table}[h]
\centering
\caption{Grouped PRPSA Results by PSA Level (Pre vs. Post)}
\label{tab:grouped-prpsa-score}
\resizebox{\columnwidth}{!}{%
\begin{tabular}{|c|c|c|c|c|c|}
\hline
\textbf{\begin{tabular}[c]{@{}c@{}}PSA Level\\ (Pre)\end{tabular}} &
\textbf{n} &
\textbf{\begin{tabular}[c]{@{}c@{}}Mean PRPSA\\ (Pre)\end{tabular}} &
\textbf{\begin{tabular}[c]{@{}c@{}}Mean PRPSA\\ (Post)\end{tabular}} &
\textbf{\begin{tabular}[c]{@{}c@{}}Mean\\ Reduction\end{tabular}} &
\textbf{\begin{tabular}[c]{@{}c@{}}Percentage\\ Reduction (\%)\end{tabular}} \\
\hline
Low      & 1 & 86.0  & 61.0  & $-25.0$ & $-29.1$\% \\ \hline
Moderate & 6 & 116.5 & 83.0  & $-33.5$ & $-28.8$\% \\ \hline
\end{tabular}
}
\end{table}

Table \ref{tab:invidual-prpsa-score-changes} presents pre and post-intervention PRPSA scores for each participant. All participants demonstrated a reduction in PRPSA scores following the use of the AI Speech Coach, with absolute reductions ranging from 19 to 45 points. Percentage reductions varied between 15.3\% and 34.4\%, indicating meaningful anxiety decreases across participants.
Six out of seven participants transitioned from moderate to low anxiety categories, while one participant (P6) remained in the moderate category but still experienced a measurable reduction. No participant exhibited an increase in anxiety, suggesting that the system did not negatively impact emotional responses to public speaking.

Table \ref{tab:grouped-prpsa-score} summarizes PRPSA changes grouped by baseline anxiety level. Participants initially classified as moderate anxiety (n = 6) exhibited an average reduction of 33.5 points, corresponding to a 28.8\%
decrease. The participant in the low anxiety category also showed a substantial reduction (29.1\%), indicating that the AI Speech Coaching application supported anxiety reduction across different initial anxiety levels, although no participants were classified as high anxiety at baseline.

\textcite{garcia2024addressing} reported statistically significant PRPSA reductions ranging from 23.3\% to 26.9\% across low, moderate, and high anxiety groups following a structured instructional intervention. The percentage reductions observed in the present study, particularly within the moderate anxiety group are comparable or slightly higher, despite differences in sample size, intervention duration, and the absence of human instructors. This comparison suggests that AI-supported, self-paced public speaking practice can
achieve anxiety reduction effects similar in magnitude to traditional structured interventions.

\section{Discussion}

The findings suggest that the context-aware AI Speech Coach was useful as a formative practice tool for academic speaking tasks. Participants showed overall improvement in public speaking performance, particularly in content clarity and organization, while the revised text and audio speeches were perceived as helpful for reflection and revision. The decrease in PRPSA scores across participants also indicates that structured, self-paced AI-supported practice may help reduce public speaking anxiety.

These findings should be interpreted cautiously. The study involved a small sample, a short intervention period, and no analysis of nonverbal aspects of speaking such as posture, gesture, and eye contact. Therefore, the results should be understood as preliminary evidence of potential rather than conclusive proof of effectiveness. Future research should test the system with larger samples, longer interventions, and multimodal analysis of speaking performance.

\section{Conclusion}

This study presented and evaluated a context-aware AI Speech Coach designed to provide contextualised feedback and revised speech examples in text and audio form. The results of the pilot study suggest that the system can support improvement in public speaking performance and may also help reduce public speaking anxiety in academic speaking tasks. In particular, the combination of context-aware feedback and model speeches appears to offer practical value for learners who need opportunities for independent, iterative practice.

At the same time, the findings remain preliminary because of the small sample size and limited study duration. For this reason, the contribution of this work is not to claim definitive effectiveness, but to demonstrate the feasibility and educational potential of a context-aware AI approach to public speaking support. Overall, the study shows that AI-based speech coaching can move beyond generic feedback by incorporating speech context and providing concrete examples for improvement, which together may make practice more meaningful, targeted, and accessible.

\section{Acknowledgment}
This work has been supported by FAST, the Finnish Software Engineering Doctoral Research Network, funded by the Ministry of Education and Culture, Finland. Also this work was funded by the European Regional Development Fund and the Regional Council of Satakunta.

\printbibliography[title={References}]

\end{document}